%% file: main.tex
\documentclass[pmlr]{jmlr}% new name PMLR (Proceedings of Machine Learning)
\usepackage{subcaption}
\usepackage{float}
\usepackage{multirow}
\usepackage{adjustbox}
\RequirePackage{graphicx}

\usepackage{longtable}% for long tables
 \usepackage{booktabs}
\usepackage[load-configurations=version-1]{siunitx} % newer version 

\usepackage{multirow}
\usepackage{soul}
\usepackage{enumitem}
\usepackage{array}
\usepackage{subcaption}

\newcolumntype{L}[1]{>{\raggedright\let\newline\\\arraybackslash\hspace{0pt}}m{#1}}
\newcolumntype{C}[1]{>{\centering\let\newline\\\arraybackslash\hspace{0pt}}m{#1}}
\newcolumntype{R}[1]{>{\raggedleft\let\newline\\\arraybackslash\hspace{0pt}}m{#1}}

\allowdisplaybreaks

\makeatletter
\def\set@curr@file#1{\def\@curr@file{#1}} %temp workaround for 2019 latex release
\makeatother
\usepackage[load-configurations=version-1]{siunitx} % newer version

\theorembodyfont{\upshape}
\theoremheaderfont{\scshape}
\theorempostheader{:}
\theoremsep{\newline}

\jmlrvolume{[VOLUME \# TBD]}
\jmlryear{2024}
\jmlrworkshop{Machine Learning for Healthcare}

\title[IMU-Video Cross-modal Self-supervision for Human Activity Recognition]{Improving Out-of-distribution Human Activity Recognition via IMU-Video Cross-modal Representation Learning}

\author{\Name{Seyyed Saeid Cheshmi}
       \Email{chesh014@umn.edu}\\ 
       \addr Department of Computer Science \& Engineering, University of Minnesota\\
       Minneapolis, MN, USA
       \AND
        \Name{Buyao Lyu}
       \Email{lyu00209@umn.edu}\\ 
       \addr Department of Mechanical Engineering, University of Minnesota\\
       Minneapolis, MN, USA
       \AND
        \Name{Thomas Lisko}
       \Email{tlisko@umn.edu}\\ 
       \addr Department of Neurosurgery, University of Minnesota\\
       Minneapolis, MN, USA
        \AND
        \Name{Rajesh Rajamani}
       \Email{rajamani@umn.edu}\\ 
       \addr Department of Mechanical Engineering, University of Minnesota\\
       Minneapolis, MN, USA
       \AND
       \Name{Robert A. McGovern}
       \Email{rmcgover@umn.edu}\\ 
       \addr Department of Neurosurgery, University of Minnesota\\
       Minneapolis, MN, USA
       \AND
       \Name{Yogatheesan Varatharajah}
       \Email{yvaratha@umn.edu}\\ 
       \addr Department of Computer Science \& Engineering, University of Minnesota\\
       Minneapolis, MN, USA
       } 

\begin{document}

\maketitle

\input{abstract}

\input{intro}

\input{related_work}

\input{method}

\input{experiment}

\input{results}

\input{discussion}

\input{acknowledgment}

\bibliography{bibl}

% \newpage
% \appendix
% \section*{Appendix A.}

% Some more details about those methods, so we can actually reproduce
% them.  After the blind review period, you could link to a repository
% for the code also.  \emph{MLHC values both rigorous evaluation as well
%   as reproduciblity.}

\end{document}

%% file: abstract.tex
\begin{abstract}
Human Activity Recognition (HAR) based on wearable inertial sensors plays a critical role in remote health monitoring. In patients with movement disorders, the ability to detect abnormal patient movements in their home environments can enable continuous optimization of treatments and help alert caretakers as needed. Machine learning approaches have been proposed for HAR tasks using Inertial Measurement Unit (IMU) data; however, most rely on application-specific labels and lack generalizability to data collected in different environments or populations. To address this limitation, we propose a new cross-modal self-supervised pretraining approach to learn representations from large-sale unlabeled IMU-video data and demonstrate improved generalizability in HAR tasks on out of distribution (OOD) IMU datasets, including a dataset collected from patients with Parkinson's disease. Specifically, our results indicate that the proposed cross-modal pretraining approach outperforms the current state-of-the-art IMU-video pretraining approach and IMU-only pretraining under zero-shot and few-shot evaluations. Broadly, our study provides evidence that in highly dynamic data modalities, such as IMU signals, cross-modal pretraining may be a useful tool to learn generalizable data representations. Our software is available at \url{https://github.com/scheshmi/IMU-Video-OOD-HAR}.

\end{abstract}

%% file: intro.tex
\section{Introduction}
% [importance of wearable based HAR and relation to PD]
Human activity recognition (HAR) using wearable inertial sensors has widespread utilities in remote health monitoring. It can provide useful information for clinical decision making by capturing abnormal movements, such as falls or near-falls, which are difficult to observe in clinical environments \citep{chen2017survey}. Recent studies have established that such techniques are particularly important for monitoring patients with movement disorders, such as Parkinson's disease (PD) \citep{cheng2017human}. Despite the availability of drug and brain-stimulation-based therapies, many PD patients continue to experience falls in their homes as those therapies primarily target abnormalities that can be measured and characterized in a clinical environment \citep{rovini2017wearable}. Compared to self-reporting and mobile-phone-based alternatives, wearable IMU sensors have been demonstrated to be both reliable and repeatable for remote HAR \citep{washabaugh2017validity}. Hence there is a critical need to develop accurate and generalizable approaches for HAR based on wearable sensing in order to most effectively classify and treat patients with movement disorders.

Machine learning (ML)-based approaches have been used very commonly to develop wearable-basd HAR models \citep{chelli2019machine, xu2017human, nouriani2023real}. However, a major challenge many ML-based models face is their lack of generalization to new data \citep{xu2021limu}. The reasons include, a) labeled activity data is typically collected in lab settings which do not reflect real-world data variability; b) there is a huge range of inertial measurement unit (IMU) sensors and wearable devices that introduce hardware- and sensing-related variability across datasets; c) differences in activity signatures between healthy participants and patients with movement disorders; and d) inter-subject differences and differences in how the wearable hardware are worn. The common approach undertaken to tackle this challenge is to develop application- and hardware-specific HAR models based on data collected from a specific device and/or environment. Although this approach addresses some of the generalization challenges, it is clearly not scalable as it requires collection of labeled data for every new device and application.

Recent availability of massive human activity datasets, such as Ego4D \citep{grauman2022ego4d}, have sparked significant interest in developing self-supervised learning (SSL)-based feature encoders that can be adapted for various downstream HAR applications with minimal finetuning \citep{saeed2019multi, xu2021limu}. Since some of these datasets also provide aligned video and/or audio segments, multimodal/cross-modal SSL approaches have also been explored \citep{moon2023imu2clip}. SSL pretraining enables feature encoders to learn useful properties of the underlying data distributions related to various activities without relying on expert labels. If the features learned during SSL pretraining are transferable to downstream applications, these feature encoders can drastically reduce the need for expert labels in downstream tasks. However, whether such transferability holds in out of distribution (OOD) settings remains unclear. Despite a few limited evaluations performed in the above SSL studies, a thorough evaluation of OOD performance of SSL-pretrained IMU encoders, particularly in challenging populations like PD patients, is currently lacking.

In this study, we propose a novel cross-modal IMU-video pretraining approach to train an IMU feature encoder and demonstrate its effectiveness in OOD HAR tasks. Specifically, we pretrain an IMU encoder using cross-modal self-supervision with $\sim$800 hours of IMU-egocentric-video pairs and evaluate its downstream performance on a publicly available activity recognition dataset ($\sim$30 hours) and a private remote health monitoring dataset of PD patients ($\sim$4 hours). We further evaluate the proposed pretraining approach by benchmarking its performance against the current state-of-the-art cross-modal pretraining approach -- IMU2CLIP \citep{moon2023imu2clip} and unimodal (IMU-only) pretraining. We also analyze the transferability of the learned features in in-distribution and OOD settings. Our results indicate that, a) IMU feature encoders pretrained on publicly available general population datasets show significant transferability to HAR tasks on OOD datasets; b) the proposed cross-modal pretraining approach outperforms the current state-of-the-art cross-modal pretraining approach in both zero-shot and few-shot evaluations; c) cross-modal SSL provides superior generalizability over unimodal SSL in low-labeled OOD settings; and d) OOD performance of the proposed cross-modal pretraining approach is at least as good as in-distribution performance when pretraining is performed on a large IMU dataset.

\subsection*{Generalizable Insights about Machine Learning in the Context of Healthcare}
ML-based human activity recognition using IMU sensors has ubiquitous applications in healthcare, and the recent availability of large-scale IMU datasets have sparked interest in developing pretrained IMU feature encoders for a range of downstream applications. In this study, we investigated the value of a novel cross-modal (IMU-video) pretraining strategy to train an IMU feature encoder and evaluated its effectiveness in out-of-distribution human activity recognition tasks. Our study provides the following generalizable insights: 
\begin{itemize}
    \item IMU encoders pretrained on publicly-available IMU datasets collected from the general population can generalize to downstream activity recognition in OOD settings, including patient populations.

    \item Cross-modal supervision provides superior generalizability over  unimodal SSL for IMU-based human activity recognition tasks in OOD settings.  
    \item Self-supervised pretraining on large-scale datasets enables learning generalizable features that transfer better to downstream tasks, even on different datasets, compared to when pretraining and evaluation are performed on the same smaller dataset.
    
\end{itemize} 
We believe that these findings can be useful for future research in wearable-based human activity recognition and pretraining of other medical feature encoders or foundation models.

%% file: related_work.tex
\section{Background \& Related Work}

Human activity recognition using wearable devices has a wide range of applications in remote health monitoring. It allows us to passively monitor activities without user input, detect falls, track rehabilitation, and supports adaptive interventions. It can be extremely useful for remotely monitoring patients with PD as injurious falls and related fractures occur frequently in PD patients \citep{palakurthi2019postural, chou2011hospitalization}. It is critical to detect abnormal movements in home environments as this information could help initiate fall prevention physical therapy programs and monitor/evaluate new treatments. Compared to self-reported diary-based \citep{hauser2006patient} and mobile-phone-based approaches \citep{lorenzi2016mobile, zhan2018using} for remote characterization of patient mobility, wearable monitoring using IMU sensors have been demonstrated to be both reliable and repeatable for measuring gait parameters remotely \citep{washabaugh2017validity}. Several previous studies have developed ML-based approaches to classify activities based on IMU sensors placed with various configurations, including sensors present in smartwatches \citep{zhang2022deep}. The supervised learning approaches studied include classical algorithms \citep{attal2015physical, xu2017human} and deep learning-based approaches \citep{yang2015deep, nouriani2023real}. However, the majority of the previous studies have relied on supervised learning using manual labels specific to a dataset. Some studies have utilized autoencoders to learn latent representations, mainly as a way to reduce dimensionality for subsequent supervised classification \citep{wang2016recognition}. 

Recent availability of large-scale IMU datasets have sparked  interest in developing SSL-based feature encoders for learning IMU representations \citep{grauman2022ego4d, fan2024emhi}. Some studies have proposed Unimodal SSL approaches to learn IMU representations: a) \citep{saeed2019multi} proposed a transformation prediction network that predicts the signal transformation as a form of self-supervision; and b) \citep{xu2021limu} proposed a BERT-like model that was pretrained using a masked reconstruction task. Since some of these datasets also provide aligned video, audio, and/or text segments, multimodal/crossmodal SSL approaches have also been explored \citep{moon2023imu2clip, gong2023mmg}. Compared to unimodal SSL, cross/multimodal SSL provides an augmented form of label-free supervision and can potentially enable learning more robust representations by aligning the semantics of different modalities \citep{alwassel2020self}. However, the cross-modal IMU-video SSL approach proposed in the above studies does not fully leverage the temporal properties of the data since the video encoder is simply an image encoder with a single input frame. In this study, we propose a novel IMU-video cross-modal SSL approach that fully leverages the temporal properties of the data by using a patch-based transformer architecture \citep{nie2022time} and spatio-temporal video encoder \citep{tong2022videomae}.

IMU data are highly dynamic and collected with diverse devices, placements, participants, and environments. These diversities cause significant domain shifts among datasets that affect the generalizability
of learned representations. However, most results reported in SSL-based studies are based on in-distribution evaluations on data collected from the general population, and very little is known about their generalization to OOD settings, particularly in patients with movement disorders. \citep{xu2021limu} report that their unimodal SSL approach showed degraded transferability on diverse datasets, however showed some transferability when controlled for the types of activities present in the datasets. \citep{saeed2019multi} evaluate their unimodal SSL approach on related datasets and show that few-shot approaches improve classification performance although zero-shot evaluations showed significant degradation. Both the cross-modal SSL studies \citep{moon2023imu2clip, gong2023mmg} have primarily focused on missing modality predictions, but downstream evaluations were performed using the same datasets used for pretraining. Studies that specifically focused on improving OOD performance in wearable-based HAR are relatively scarce. \citep{roy2023classifying} developed a deep time ensemble-approach that utilized temporal randomness during supervised pretraining to improve model robustness and showed improvements in classifying activities that were not seen during pretraining. To our knowledge, no prior studies have evaluated the value of a novel cross-modal SSL feature encoder in improving OOD performance on substantially different population of participants.

%% file: method.tex
\section{Methods}
\label{sec:methods}

\begin{figure}[h]
   \centering 
   \includegraphics[width=0.99\textwidth]{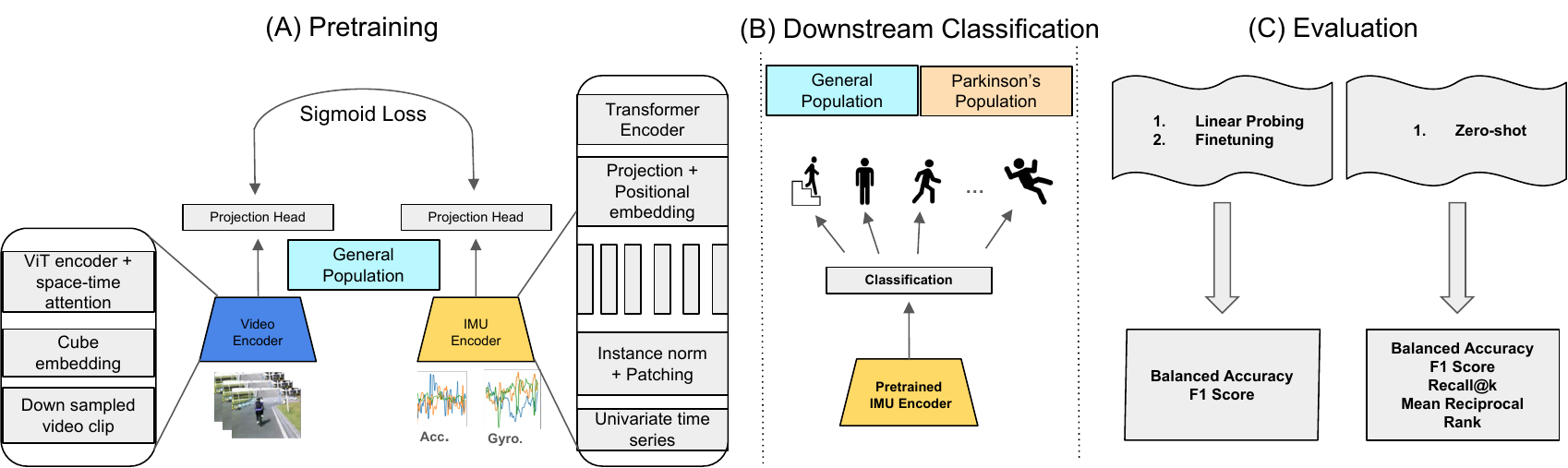} 
   \caption{Overall workflow of our approach. (A) illustrates our pretraining approach, including the IMU and video encoders. Subsequently, we use the pretrained IMU encoder for downstream classification of activities (B). We then perform downstream evaluations with zero-shot, linear probing, and finetuning, and use several metrics for comparison (C).}
   \label{fig:workflow} 
\end{figure} 

The overall workflow of our approach is illustrated in Figure \ref{fig:workflow}. Our workflow comprises, a) a cross-modal pretraining approach to align IMU and video representations using a large paired IMU-video dataset from the general population; b) downstream classifications of human activities in both general and Parkinson's populations; and c) evaluations performed with zero-shot, linear probing, and finetuning and comparisons with various baselines. We describe each of these steps in the following sections.

\subsection{Pretraining}
\label{subsec:pretraining}

We develop a cross-modal pretraining approach to align the representation of a given IMU window to the representation of the paired video segment. To this end, we employ a contrastive learning approach between IMU and video encoders to bring the paired IMU and video frame representations closer in the shared embedding space, while pushing apart the representations of unrelated IMU-video pairs. We describe the two encoders below. % are trained to make paired IMU and video embeddings similar. 

\noindent\textbf{IMU encoder}: We utilize PatchTST (illustrated in Figure \ref{fig:workflow}A) as the IMU encoder $f(\cdot)$ \citep{nie2022time}. PatchTST treats multivariate time series as multiple univariate time series and employs the vanilla Transformer \citep{vaswani2017attention} encoder as its core architecture. We briefly describe the components of this architecture below. Readers are encouraged to read the original article for a more detailed description. 

\begin{itemize}[leftmargin=*]
  \item \ul{Channel independence}: Given a multivariate time series with a context length of $L$: $(x_1, x_2, \dots, x_L), \text{ where } x_t \in \mathbb{R}^{M}$, each of the $M$ channels is treated independently, which helps the model capture unique temporal patterns within individual channels. The multivariate input is thus separated into $M$ independent univariate series $(x^{(i)}_1, x^{(i)}_2, \dots, x^{(i)}_L)$. Here $i\in \{1,\dots,M\}$ represents a channel in the multivariate input timeseries.  
  \item \ul{Patching}: Furthermore, instead of using single time points, each univariate series is segmented into patches to preserve local semantics and significantly reduce computational complexity. For a given univariate series $x^{(i)}$, it is segmented into patches of length $P$ with stride $S$. The resulting patched series is $x_p^{(i)} \in \mathbb{R}^{P \times N}, \text{ where } N = \left\lfloor \frac{L - P}{S} \right\rfloor + 2$.
  \item \ul{Temporal modeling}: Each patch $x_p^{(i)}$ is then projected into a latent Transformer embedding space. A trainable linear projection and positional embedding are applied to track the temporal order of patches. In the multi-head self-attention mechanism (with $H$ heads), the latent representation is transformed into query, key, and value matrices for each head. Then the output is computed using scaled dot-product attention. % The multi-head attention block also includes batch normalization, a feed-forward network, and residual connections. 
  \item \ul{Output}: We use $z^{(i)}$ to denote the output representation from the Transformer encoder for channel $i$. Since we need an embedding for the given context length to further align with corresponding video frames, we add an additional token (\textit{CLS token}) among patches, which learns a general representation of the given time-series context length. Thus, the final representation has the shape $D \times (N+1)$, where $D$ is the embedding dimensionality.
\end{itemize}

\noindent\textbf{Video encoder}: We utilize VideoMAE (illustrated in Figure \ref{fig:workflow}A) pretrained on an egocentric video dataset (SSV2, \cite{goyal2017something}) as the video encoder \citep{tong2022videomae}. Due to the significant redundancy present in videos, we can significantly downsample the video in  time without losing important information. VideoMAE utilizes vanilla vision transformer architecture with joint space-time attention and carries out a masked autoencoding procedure to learn effective spatiotemporal representations \cite{dosovitskiy2020image}. It takes the downsampled frames as inputs and uses cube embedding to obtain video tokens. Then, tube masking is applied with a high masking ratio to perform pretraining.

\noindent\textbf{Aligning IMU and video encoders}: Our goal here is to align the embedding of the [CLS] token from the IMU encoder and the average embedding from the video encoder. We use a projection head on top of IMU and video embeddings to bring them into another space with the same dimensionality. We then use a sigmoid-based contrastive loss, introduced in \citep{zhai2023sigmoid}, to align the projected representations of IMU and video encoders. We chose this loss due to limited hardware budget and since this loss has been shown to be effective for contrastive training small batch sizes. Suppose that we denote a batch of training windows as $\mathcal{B} = \{(I_1, V_1), (I_2, V_2), \dots\}$ and use $f(\cdot)$ to denote the IMU encoder and $g(\cdot)$ to denote the video encoder. The loss function is defined as:

\[
- \frac{1}{|\mathcal{B}|} \sum_{i=1}^{|\mathcal{B}|} \sum_{j=1}^{|\mathcal{B}|} 
\log { \frac{1}{1 + e^{z_{ij} \left( -t\, \mathbf{i}_i \cdot \mathbf{v}_j + b \right)} }}
\label{loss_function}
\]
% \(\mathbf{i}_i\) and \(\mathbf{v}_i\) similar,
where $\mathbf{i}_i = \frac{f(\cdot)}{\|f(\cdot)\|_2} \quad \text{and} \quad \mathbf{v}_i = \frac{g(\cdot)}{\|g(\cdot)\|_2}.$ This loss function processes each pair independently and converts the problem into binary classification on the set of all pairs. Where, \( z_{ij} \) is a label for a given pair which is 1 if \( (I_i, V_i) \) and -1 if \( (I_i, V_j) \) with \( j \neq i \). The heavy imbalance comes from a large number of negative pairs, causing the model to take large optimization steps initially in an effort to compensate for this imbalance. To address this issue, an additional learnable bias term \( b \), and the temperature \( t \) are added. We initialize \( t \) and \( b \) to $e^{\log 10}$ and \(-10\), respectively, as mentioned in the original article. During pretraining, all parameters in IMU and video encoders are updated. 

\subsection{Datasets}
\label{subsec:datasets}
\begin{figure}[h]
   \centering 
   \includegraphics[width=0.95\textwidth]{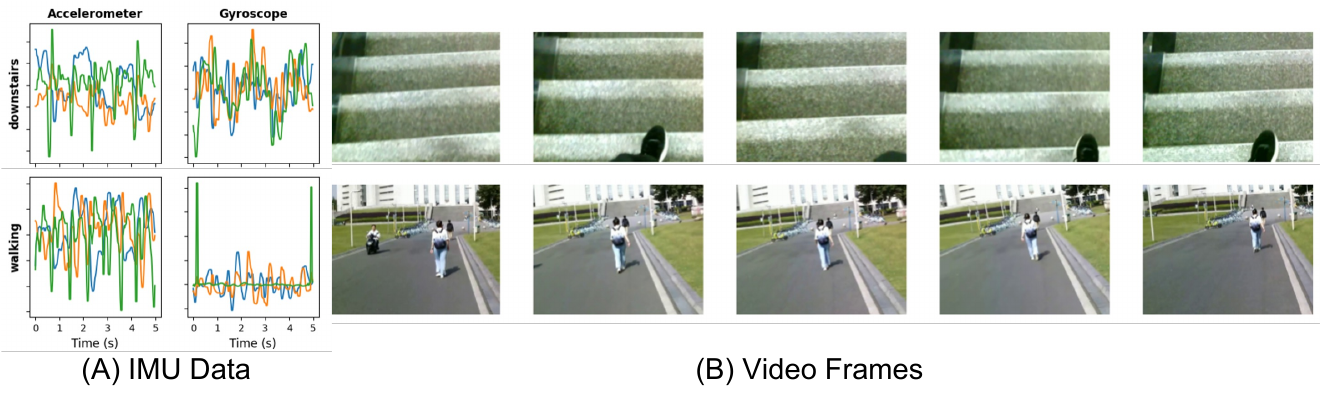} 
   \caption{Sample paired IMU and video data segments from the MMEA dataset.}
   \label{fig:sample_imu} 
 \end{figure}

In this study, we make use of the following datasets (summarized in Table \ref{tab:datasets}): a) Ego4D \citep{grauman2022ego4d} - a large dataset containing paired IMU and egocentric videos from a general population; b) MMEA \citep{xu2023towards} - a relatively smaller activity recognition dataset containing paired IMU and egocentric videos from a general population; and c) PD - a smaller dataset containing home-based IMU measurements from $4$ patients with movement disorders. In all the above datasets, we utilized IMU data including three accelerometers ($\textrm{x,y,z}$) and three gyroscopes ($\textrm{x,y,z}$), resulting in a 6-channel timeseries. Furthermore, to be consistent with the location of the IMU sensors, we utilized either head-worn or chest-worn IMU measurements. Figure \ref{fig:sample_imu} shows sample IMU and paired video frames for two activities (walking and going downstairs) from the MMEA dataset.

\begin{table}[h]
  \centering
  \caption{Summary of datasets used in our study.}
  \label{tab:datasets}
  \footnotesize
  \renewcommand{\arraystretch}{1.2}
  \begin{adjustbox}{width=\textwidth}
    \begin{tabular}{
      >{\centering\arraybackslash}p{2.2cm}  % Dataset
      >{\centering\arraybackslash}p{2.5cm}  % Modalities
      >{\centering\arraybackslash}p{2.2cm}  % Population
      >{\centering\arraybackslash}p{1.6cm}  % Length
      >{\centering\arraybackslash}p{1.6cm}  % # Classes
      >{\centering\arraybackslash}p{2.2cm}  % # Windows
      >{\centering\arraybackslash}p{2.2cm}  % Avg. windows/class
    }
      \toprule
      \textbf{Dataset} & \textbf{Modalities} & \textbf{Population} & \textbf{Length (h)} & \textbf{\# Classes} & \textbf{\# 5-sec windows} & \textbf{Avg. windows/class} \\
      \midrule
      Ego4D & Video / IMU     & General     & 815   & -  & $\sim$587K & -   \\
      MMEA  & Video / IMU     & General     & 30.4  & 32 & $\sim$22K  & 684 \\
      PD    & IMU             & PD patients & 4     & 5  & $\sim$3K   & 628 \\
      \bottomrule
    \end{tabular}
  \end{adjustbox}
\end{table}

\noindent\textbf{Ego4D}: Ego4D is a large, publicly available dataset of egocentric videos \citep{grauman2022ego4d}. It spans roughly 3,670 hours of unscripted video from 931 unique camera-wearers across 74 global locations, spanning a wide age range (from under 18 to over 80), with a near-balanced gender distribution (55\% male, 45\% female). The contributors used head-mounted cameras to capture everyday scenarios such as cooking, social gatherings, commuting, playing sports, and more. The data covers not just RGB streams but often includes audio, multiple camera views, IMU signals, and even 3D scans of the environment for some subsets. In this work, we utilize $\sim$815 hours of IMU-video pairs from this dataset.

\noindent\textbf{MMEA}: This dataset was specifically designed for continual activity recognition and contains around 30.4 hours of video paired with synchronized IMU data \citep{xu2023towards}. Data was collected from 10 participants, each performing 32 different daily activities (e.g., walking, cooking, washing dishes) while wearing custom-developed smart glasses that embed both a first-person camera and inertial sensors. Each activity was sampled for $\sim$200 times, each for $\sim$15-18s. The video is recorded at 25 frames per seond with a resolution of 640x480, and the IMU data is sampled at the same rate to maintain strict synchronization. % Figure \ref{fig:sample_imu} shows a sample data from this dataset.

\noindent\textbf{PD}: IMU data from four patients with Parkinson's disease who wore a custom built IMU sensor-kit for several days to record their day-to-day activities. Although the patients wore around five sensor kits at different body locations, we utilized measurements from a chest-worn sensor kit to be consistent with other datasets. The data were labeled into five activities, walking, turning, bending, standing, and sitting, with each activity comprising 108, 10, 30, 32, and 61 minutes of IMU data, respectively.

\noindent\textbf{Preprocessing}: We apply the same preprocessing steps to all three datasets before any training or evaluation. First, we convert the IMU data to a 50 Hz sampling rate by either upsampling or downsampling, depending on the original frequency. Next, we apply median filtering with a kernel size of 5 for noise removal, and finally we normalize the signals using z-score normalization. In all settings, once preprocessing is complete, we extract 5 seconds windows (which corresponds to 250 timestamps at 50 Hz) that serve as inputs to the model. For cross-modal models, we also extract aligned video segments. After extracting the corresponding 5-second video segment, we divide all the frames in that segment into 10 equal chunks and randomly select one frame from each chunk, resulting in 10 frames representing a 5-second segment. This frame-selection strategy helps capture the overall flow of the action within a given window of video segment.

\subsection{Evaluation}
\label{subsec:eval}

We evaluate the pretrained encoder in several ways, including zero-shot classification, linear probing, and full-model finetuning, for classification of activities in downstream tasks. All evaluations are performed on OOD datasets, MMEA and PD.  

\noindent\textbf{Zero-shot evaluation}: We first select prototype video segments for each action from both downstream datasets. We then encode these prototype videos using our pretrained video encoder, apply a nearest neighbor classifier in the embedding space with all the IMU embeddings resulting from downstream datasets, and subsequently classify each IMU window based on the activity class of the closest video embedding.

\noindent\textbf{Few-shot evaluation}: We evaluate few-shot performance by further tuning the pretrained encoder via linear probing \& full-model finetuning. We perform linear probing by adding a linear classification layer on top of the pretrained encoder. For full-model finetuning, we again add a linear classification layer on top of the pretrained IMU encoder; however, we fine-tune both the encoder and the linear layer weights.

\noindent\textbf{Evaluation metrics}: Our evaluations utilize the following metrics listed in Table \ref{tab:eval_metrics}. Note that the metrics Recall@k and Mean Reciprocal Rank are utilized only in zero-shot evaluations and all evaluations utilize balanced accuracy measure to compare performances.

\begin{table}[h]
  \centering
  \caption{Evaluation metrics used in our study.}
  \label{tab:eval_metrics}
  \footnotesize
  \renewcommand{\arraystretch}{1.2}
  \begin{adjustbox}{width=\textwidth}
    \begin{tabular}{
      >{\centering\arraybackslash}p{2.5cm}  % Metric
      >{\centering\arraybackslash}p{3.2cm}  % Definition
      >{\raggedright\arraybackslash}p{3.8cm}  % Notations
      >{\raggedright\arraybackslash}p{6.5cm}  % Description
    }
      \toprule
      \textbf{Metric} & \textbf{Definition} & \textbf{Notations} & \textbf{Description} \\
      \midrule

      Balanced Accuracy & $\frac{1}{C} \sum_{i=1}^{C} 
      \frac{\text{TP}_i}{\text{TP}_i + \text{FN}_i}$ & $C$ - \# classes & Measures performance in the presence of class imbalance setting by averaging the recall across all classes.\\

      \midrule
      Recall@k & $\frac{1}{N}\sum_{i=1}^{N}\mathbf 1_i$ & $\mathbf 1_i$ - indicates whether the ground‑truth class appears in the top-$k$ predictions & For a given IMU sample, we compute similarity scores with respect to the prototype video embeddings and sort these scores, which yields a ranking of the classes. A higher \(\operatorname{Recall@}k\) indicates that the model retrieves the correct class within the first \(k\) positions.\\

      \midrule
      Mean Reciprocal Rank (MRR) & $\frac{1}{N}\sum_{i=1}^{N}\frac{1}{\operatorname{rank}_i}$ & \(\operatorname{rank}_i\) denotes the 1‑based position of the ground‑truth class in the sorted list for query $i$. & A higher MRR indicates that the model tends to place the correct class closer to the top of the ranking.\\

      \bottomrule
    \end{tabular}
  \end{adjustbox}
\end{table}

%% file: experiment.tex
\section{Experimental Setup}
\label{sec:experiments}

In our experiments, we perform the following analyses focusing on OOD generalization. 

\begin{itemize}[leftmargin=*]
  \item \ul{Unimodal (IMU-only) pretraining vs cross-modal (IMU-video) pretraining}: we first evaluate whether cross-modal SSL outperforms unimodal SSL in OOD performance. To that end, in addition to the cross-modal pretraining approach described in Section \ref{subsec:pretraining}, we also pretrain a unimodal IMU-only encoder using a masked reconstruction task with the same PatchTST encoder. We separately pretrain unimodal and cross-modal SSL encoders on Ego4D and MMEA datasets, and evaluate few-shot performances on OOD HAR tasks in the datasets that were not used during pretraining.
  \item \ul{Proposed cross-modal pretraining vs state-of-the-art (SOTA)}: second, we pretrain cross-modal (IMU-video) SSL encoders using the SOTA IMU2CLIP \citep{moon2023imu2clip} and the proposed approach, and evaluate their zero-shot and few-shot performances in the OOD dataset of PD patients. We perform these comparisons by pretraining both encoders on Ego4D and MMEA and evaluating on the datasets that were not used during pretraining. 
  \item \ul{Cross-modal pretraining in OOD settings vs in-distribution settings}: third, we evaluate the value of cross-modal SSL pretraining in OOD settings by comparing the transferability of a cross-modal pretrained encoder in OOD and in-distribution settings. We perform this comparison by pretraining the proposed cross-modal encoder using the data from 27 of the 32 classes present in the MMEA dataset, and evaluating the pretrained encoder in the rest of the classes in MMEA and in the PD dataset.
\end{itemize}

\noindent\textbf{Training details}: All experiments were performed using a Ubuntu workstation with two Nvidia RTX 4090 graphics processing units. During cross-modal pretraining, we divide the preprocessed IMU timeseries into 5-second non-overlapping windows and input them to IMU encoder with a context length of 250, and with patch size and stride of 16. We then select 10 frames from each corresponding video segment and input them into the video encoder. We then train the contrastive alignment objective for 50 epochs with a batch size of 32, using the AdamW optimizer \citep{loshchilov2017decoupled}, a learning rate of 1e-4, and a cosine annealing learning rate scheduler. For IMU-only pretraining, we divide each preprocessed IMU time series into 5-second non-overlapping patches, then randomly mask entire patches, and train the model using a reconstruction loss to predict missing patches. We set the context length to 250, patch size to 16, stride to 16, and the masking ratio to 40\%. Then the model is trained for 100 epochs. We also compare these encoders with a fully-supervised baseline using the same experimental approach as the unimodal finetuning experiment except that the encoder was trained from scratch in a supervised fashion.

During few-shot linear probing, we add a linear layer to the pretrained IMU encoder, freeze the encoder weights, and only update the final linear layer weights. This training is done using the AdamW optimizer with a learning rate of \(1 \times 10^{-3}\) for 25 epochs. During full-model finetuning, we update both the encoder and the linear layer weights. Here we use a learning rate of \(1 \times 10^{-6}\) for the pre-trained encoder and \(1 \times 10^{-3}\) for the classification layer, and the remaining training configuration is the same as the linear probing.

\noindent\textbf{Evaluation details}: For evaluating few-shot performances, we perform linear probing and full-model finetuning five times using randomly selected subsets of 10, 20, 50, and 100 samples per class and test classification performance using held-out 20 samples per class. We then compute the balanced accuracy for each run on the held-out set and report average results. For zero-shot evaluation, we bootstrap 80\% of the samples from each class (from either MMEA or PD data) and compute balanced ccuracy, F1-score, MRR, R@1, and R@3. We repeat this process five times and report averaged results.

%% file: results.tex
\section{Results}

Our results are summarized in Figure \ref{fig:result_figure}, Table \ref{tab:fewshot}, and Table \ref{tab:zeroshot}. In the following, we describe the findings for each of the experiments listed in Section \ref{sec:experiments}.

\subsection{Unimodal pretraining vs cross-modal pretraining}

\begin{figure}[h]
   \centering 
    \includegraphics[trim={1.5cm 1.2cm 1.5cm 0.2cm},clip,width=\textwidth]{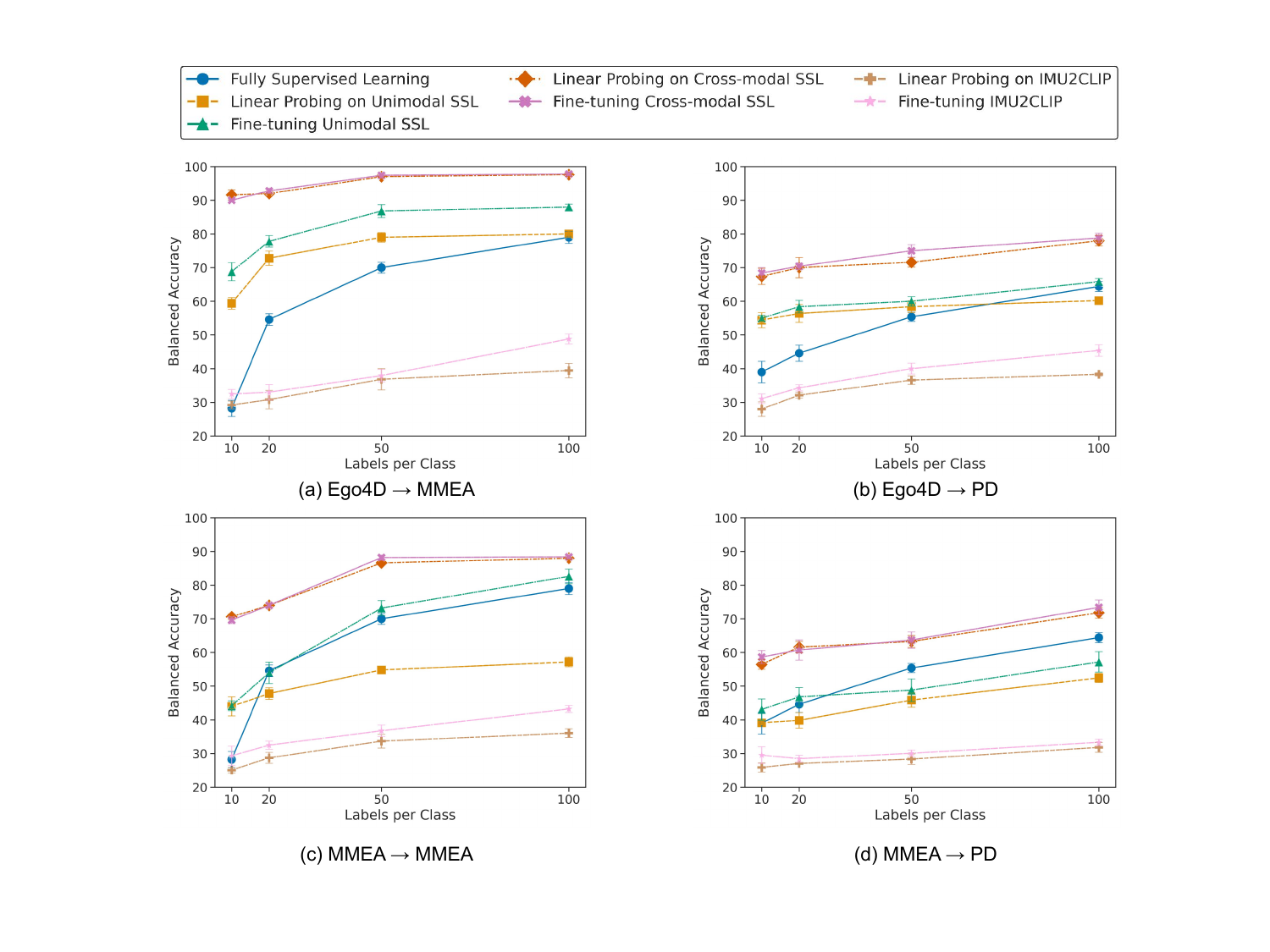}
    \caption{Few-shot performance of unimodal (IMU-only) and cross-modal (IMU-video) pretrained encoders and comparison with fully-supervised learning. Figure captions indicate the pretraining dataset and the downstream evaluation dataset, respectively.}
   \label{fig:result_figure}
 \end{figure}

Table \ref{tab:fewshot} and Figure \ref{fig:result_figure} indicate that cross-modal (IMU-video) pretraining significantly outperforms unimodal (IMU-only) pretraining, regardless of whether the pretraining is performed on the large Ego4D dataset or on the smaller MMEA dataset. For example, the cross-modal pretrained IMU encoder on Ego4D is able to reach $>$90\% balanced accuracy in few-shot classification of MMEA dataset using only 10 labels per class, either through linear probing or fine-tuning. In contrast, with the same number of labels, the best balanced accuracy achievable using a unimodal encoder pretrained on Ego4D is only 68.83\%. A similar pattern is observed when the model is pretrained on MMEA and evaluated on the PD data. For instance, by using only 10 labeled samples from the PD dataset, the cross-modal pretrained encoder achieves $\sim$17\% and $\sim$15\% improvements in balanced accuracy with linear probing and full-model finetuning, respectively, compared to the unimodal pretrained encoder. Results also indicate that increasing the number of labels per class further improves the balanced accuracy regardless of the few-shot paradigm.

\subsection{Proposed cross-modal pretraining vs SOTA IMU2CLIP}

We find that the proposed cross-modal pretraining approach outperforms the state-of-the-art IMU2CLIP pretraining approach in both zero-shot and few-shot evaluations in OOD settings. Table \ref{tab:zeroshot} provides a comparison between the two approaches in zero-shot OOD evaluations. We find that the zero-shot performance of the proposed pretraining approach is superior to IMU2CLIP based on all the metrics of comparison. The proposed approach achieves $\sim$5\% and $\sim$12\% improvements in balanced accuracy under zero-shot evaluations on PD and MMEA datasets, when the encoder was pretrained on Ego4D. Our approach also achieved $\sim$11\% improvement in an in-distribution evaluation where both pretraining and finetuning were performed on the MMEA dataset. Table \ref{tab:fewshot} provides a comparison between the two approaches in few-shot OOD evaluations. We find that the highest balanced accuracy achieved by IMU2CLIP after fine-tuning using 100 labels per class on MMEA is 48.80\%. In contrast, with the same number of labels used for few-shot evaluation, the proposed method achieves 97.81\% balanced accuracy. A similar pattern is observed on the PD patient dataset, where fine-tuning our pretrained encoder yields $>$33\% higher balanced accuracy compared to fine-tuning the IMU2CLIP-pretrained encoder using 100 labels.

\begin{table}[t]
  \centering
  \caption{Few-shot performance of unimodal (IMU-only) and cross-modal (IMU-video) pretrained encoders in classifying activities.}
  \label{tab:fewshot}
  \footnotesize
  \begin{adjustbox}{width=\textwidth}
    \begin{tabular}{
      >{\centering\arraybackslash}p{2.8cm}  % Pretraining Dataset
      >{\centering\arraybackslash}p{2.8cm}  % Approach
      >{\centering\arraybackslash}p{2.2cm}  % Experiment + # labels
      >{\centering\arraybackslash}p{2cm}    % LP MMEA
      >{\centering\arraybackslash}p{2cm}    % LP PD
      >{\centering\arraybackslash}p{2cm}    % FT MMEA
      >{\centering\arraybackslash}p{2cm}    % FT PD
    }
      \toprule
      \textbf{Pretraining Dataset} & \textbf{Approach} & 
      \begin{tabular}[c]{@{}c@{}}\textbf{Experiment} \\[-0.2ex] \textbf{\# labels} \end{tabular} &
      \multicolumn{2}{c}{\textbf{Linear Probing}} & 
      \multicolumn{2}{c}{\textbf{Finetuning}} \\
      \cmidrule(lr){4-5} \cmidrule(lr){6-7}
      & & &\textbf{MMEA} & \textbf{PD} & \textbf{MMEA} & \textbf{PD} \\
    \midrule

    \multirow{12}{*}{Ego4D} 
    & \multirow{4}{*}{IMU-only SSL} 
      & 10  & 59.42 ± 1.74 & 54.42 ± 2.24 & 68.83 ± 2.71 & 55.94 ± 0.89 \\
    &   & 20  & 72.83 ± 2.13 & 56.41 ± 2.65 & 77.86 ± 1.72 & 58.46 ± 1.85 \\
    &   & 50  & 79.73 ± 1.41 & 58.43 ± 1.39 & 86.89 ± 1.93 & 60.19 ± 1.41 \\
    &   & 100 & 80.22 ± 1.00 & 60.23 ± 0.58 & 88.32 ± 0.89 & 65.82 ± 0.97 \\
    \cmidrule(lr){2-7}

    & \multirow{4}{*}{
  \centering
  \begin{tabular}[c]{@{}c@{}}
    IMU2CLIP \\
    \textcolor{blue}{Moon et al. (2023)}
  \end{tabular}
}
      & 10  & 29.22 ± 1.16 & 28.00 ± 2.16 & 32.46 ± 1.35 & 31.08 ± 1.41 \\
    &   & 20  & 30.80 ± 2.74 & 32.12 ± 0.81 & 33.07 ± 2.28 & 34.33 ± 0.94 \\
    &   & 50  & 36.84 ± 3.17 & 36.66 ± 1.24 & 37.93 ± 1.83 & 40.18 ± 1.63 \\
    &   & 100 & 39.43 ± 2.15 & 38.31 ± 0.47 & 48.80 ± 1.46 & 45.40 ± 1.73 \\
    \cmidrule(lr){2-7}

    & \multirow{4}{*}{\textbf{Ours**}} 
      & 10  & 91.62 ± 1.49 & 67.40 ± 2.36 & 90.28 ± 0.02 & 68.49 ± 1.62 \\
    &   & 20  & 92.34 ± 0.05 & 70.33 ± 3.21 & 92.82 ± 0.74 & 70.40 ± 1.06 \\
    &   & 50  & 97.12 ± 0.03 & 71.61 ± 1.47 & 97.46 ± 0.82 & 75.22 ± 1.88 \\
    &   & 100 & 97.81 ± 0.48 & 78.27 ± 0.59 & 97.81 ± 0.40 & 78.86 ± 1.51 \\
    
    \midrule

    \multirow{12}{*}{MMEA} 
    & \multirow{4}{*}{IMU-only SSL} 
      & 10  & 44.66 ± 2.82 & 39.11 ± 0.63 & 44.23 ± 1.46 & 43.06 ± 3.16 \\
    &   & 20  & 47.84 ± 1.72 & 39.84 ± 2.31 & 54.31 ± 3.16 & 46.82 ± 2.73 \\
    &   & 50  & 54.87 ± 0.84 & 45.87 ± 2.13 & 73.24 ± 2.22 & 48.81 ± 3.34 \\
    &   & 100 & 57.26 ± 1.46 & 52.46 ± 1.35 & 82.64 ± 2.15 & 57.19 ± 3.03 \\
    \cmidrule(lr){2-7}

    & \multirow{4}{*}{
  \centering
  \begin{tabular}[c]{@{}c@{}}
    IMU2CLIP \\
    \textcolor{blue}{Moon et al. (2023)}
  \end{tabular}
}
      & 10  & 25.06 ± 0.81 & 25.88 ± 1.37 & 29.33 ± 2.86 & 29.51 ± 2.50 \\
    &   & 20  & 28.71 ± 1.64 & 27.04 ± 0.27 & 32.47 ± 1.19 & 28.45 ± 1.05 \\
    &   & 50  & 33.66 ± 2.05 & 28.33 ± 1.56 & 36.75 ± 1.69 & 30.08 ± 0.96 \\
    &   & 100 & 36.06 ± 1.26 & 31.81 ± 1.42 & 43.23 ± 1.03 & 33.31 ± 0.93 \\
    \cmidrule(lr){2-7}

    & \multirow{4}{*}{\textbf{Ours**}} 
      & 10  & 70.64 ± 0.82 & 56.83 ± 1.46 & 69.62 ± 1.02 & 58.63 ± 1.93 \\
    &   & 20  & 74.36 ± 0.63 & 61.60 ± 1.69 & 74.32 ± 0.53 & 60.73 ± 2.99 \\
    &   & 50  & 86.62 ± 0.48 & 63.27 ± 1.75 & 88.26 ± 0.64 & 63.68 ± 2.41 \\
    &   & 100 & 88.06 ± 0.23 & 71.80 ± 1.57 & 88.44 ± 0.48 & 73.40 ± 2.19 \\
    
        \bottomrule
    \end{tabular}
   \end{adjustbox}
\end{table}

\begin{table}[t]
  \centering
  \caption{Zero-shot performance of our proposed cross-modal encoder and SOTA baseline.}
  \label{tab:zeroshot}
  \footnotesize
  \renewcommand{\arraystretch}{1.2}
  \begin{adjustbox}{width=\textwidth}
    \begin{tabular}{
      >{\centering\arraybackslash}p{2.3cm}  % Metric
      >{\centering\arraybackslash}p{2.5cm}  % IMU2CLIP 1
      >{\centering\arraybackslash}p{2.5cm}  % IMU2CLIP 2
      >{\centering\arraybackslash}p{2.5cm}  % IMU2CLIP 3
      >{\centering\arraybackslash}p{2.5cm}  % Ours 1
      >{\centering\arraybackslash}p{2.5cm}  % Ours 2
      >{\centering\arraybackslash}p{2.5cm}  % Ours 3
    }
      \toprule
      \textbf{Metrics} & \multicolumn{3}{c}{\textbf{IMU2CLIP}} & \multicolumn{3}{c}{\textbf{Ours**}} \\
      \cmidrule(lr){2-4} \cmidrule(lr){5-7}
      & Ego4D $\rightarrow$ PD & Ego4D $\rightarrow$ MMEA & MMEA $\rightarrow$ MMEA
      & Ego4D $\rightarrow$ PD & Ego4D $\rightarrow$ MMEA & MMEA $\rightarrow$ MMEA \\
      \midrule

      B. Acc. & 19.530 ± 0.063 & 20.562 ± 0.404 & 22.124 ± 1.245 & 24.137 ± 0.940 & 32.852 ± 1.166 & 33.851 ± 1.210 \\
      F1      & 0.098 ± 0.007  & 0.117 ± 0.097  & 0.247 ± 0.034  & 0.238 ± 0.013  & 0.298 ± 0.017  & 0.334 ± 0.031 \\
      MRR     & 0.463 ± 0.041  & 0.440 ± 0.014  & 0.462 ± 0.024  & 0.535 ± 0.067  & 0.549 ± 0.021  & 0.563 ± 0.014 \\
      R@1     & 0.209 ± 0.006  & 0.181 ± 0.015  & 0.206 ± 0.023  & 0.285 ± 0.015  & 0.310 ± 0.036  & 0.309 ± 0.053 \\
      R@3     & 0.721 ± 0.011  & 0.572 ± 0.073  & 0.606 ± 0.051  & 0.731 ± 0.009  & 0.677 ± 0.010  & 0.773 ± 0.029 \\

      \bottomrule
    \end{tabular}
  \end{adjustbox}
\end{table}

\subsection{Out-of-distribution vs in-distribution performance on MMEA}

Here we compare the transferability of representations learned by the proposed pretraining approach between in-distribution and OOD settings. Specifically, we evaluate the following scenarios: 1) pretraining on MMEA followed by few-shot evaluation on MMEA, and (2) pretraining on Ego4D followed by few-shot evaluation on MMEA.
Table \ref{tab:fewshot} and Figure ~\ref{fig:result_figure}c indicate that when pretraining and evaluation are both performed on MMEA, we achieve $\sim$88\% balanced accuracy with 100 labeled samples using either linear probing or finetuning. However, when pretraining is performed on Ego4D, we achieve $>$90\% balanced accuracy on MMEA — even when using only 10 labeled samples per class — through linear probing or finetuning. The transferability of the learned representations can also be observed in the zero-shot evaluation of our proposed approach. As shown in Table \ref{tab:zeroshot}, regardless of whether the encoder was pretrained on Ego4D or MMEA, the zero-shot performance on MMEA is similar — 32.85\% and 33.85\%, respectively.

\subsection{Contributions of the various architectural components}
We performed additional experiments to quantify the contribution of each component in the proposed framework. To that end, we performed a component-level analysis by replacing the video encoder, IMU encoder, and the loss function with their counterparts from the SOTA IMU2CLIP model. We pretrained each variant on the Ego4D dataset and evaluated using linear probing and fine-tuning on the MMEA and PD datasets with 100 labeled samples.
The results (shown in Table \ref{tab:ablation}) indicate that utilizing the the VideoMAE encoder instead of the image-based ViT-CLIP encoder provided very significant gains (37-40\% on MMEA and 22-24\% on PD) in dwonstream performance, highlighting the importance of capturing temporal dynamics for action recognition. 
Additionally, the IMU encoder (i.e., PatchTST) also provided significant gains (21-25\% on MMEA and 11-13\% on PD) over the (1D-CNN + RNN) encoder utilized in IMU2CLIP. 
Lastly, substituting the softmax layer with a sigmoid activation provided further gains (6-7\% on MMEA and 4-5\% on PD), suggesting that the loss sigmoid is more suitable in this context.
These observations were consistent across both linear probing and fine-tuning evaluations, and in both downstream datasets.

\begin{table}[ht]
  \centering
  \caption{Quantifying the contributions of the architectural components by replacing individual components with their counterparts from the SOTA IMU2CLIP model. Evaluations included linear probing and finetuning on MMEA and PD datasets using 100 labels.}
  \label{tab:ablation}
  \footnotesize
  \begin{adjustbox}{width=\textwidth}
    \begin{tabular}{
      >{\centering\arraybackslash}p{5.2cm}  
      >{\centering\arraybackslash}p{2.1cm}  
      >{\centering\arraybackslash}p{2.1cm}  
      >{\centering\arraybackslash}p{2.1cm}  
      >{\centering\arraybackslash}p{2.1cm}  
    }
      \toprule
      \textbf{Method} &
      \multicolumn{2}{c}{\textbf{Linear Probing}} & 
      \multicolumn{2}{c}{\textbf{Fine-Tuning}} \\
      \cmidrule(lr){2-3} \cmidrule(lr){4-5}
      (All pretrained on Ego4D) & \textbf{MMEA} & \textbf{PD} & \textbf{MMEA} & \textbf{PD} \\
      \midrule

      IMU2CLIP & 39.43 ± 2.15 & 38.31 ± 0.47 & 48.80 ± 1.46 & 45.40 ± 1.73 \\
      Ours w/ Softmax Loss & 90.89 ± 1.46 & 72.98 ± 1.31 & 91.62 ± 1.01 & 74.83 ± 1.60 \\
      Ours w/  (1D-CNN + RNN)  & 72.34 ± 0.48 & 65.63 ± 1.58 & 76.42 ± 0.11 & 67.38 ± 1.40 \\
      Ours w/ ViT-CLIP  & 57.06 ± 1.03 & 54.17 ± 1.08 & 60.97 ± 1.37 & 56.32 ± 1.21 \\
      Ours & 97.81 ± 0.48 & 78.27 ± 0.59 & 97.81 ± 0.40 & 78.86 ± 1.51 \\

      \bottomrule
    \end{tabular}
  \end{adjustbox}
\end{table}

%% file: discussion.tex
\section{Discussion}
In this work, we proposed a new cross-modal self-supervised pretraining approach based on paired IMU and video data to improve out-of-distribution human activity recognition. We evaluated our method on one public dataset and one private dataset of Parkinson's disease patients, covering two different populations, general and clinical. Furthermore, we perform our evaluations using zero-shot, few-shot linear probing, and few-shot fine-tuning to showcase the label efficiency of our approach. Our results show that with less than 9 minutes (100 5-second windows) of IMU recording of an action, which is relatively easy to collect, we can achieve reasonable performance in recognizing that action. 

We performed experiments to evaluate, a) the benefits of cross-modal pretraining over unimodal pretraining in OOD generalization; b) a comparison of the proposed cross-modal pretraining approach with the current state-of-the-art on OOD activity recognition tasks; and c) a comparison between in-distribution and OOD performance. Our findings highlight the following. First, cross-modal (IMU-video) pretraining outperforms unimodal (IMU-only) pretraining for the task of human activity recognition using wearable IMU sensor data. We believe that video-based cross-modal supervision provides a type of regularization during pretraining that steers the learned representations in meaningful directions associated with various activities. Second, the proposed cross-modal pretraining outperforms current state-of-the-art (IMU2CLIP) pretraining approach, highlighting the quality of the representations learned by the proposed cross-modal pretraining method. We believe that this is due to the specific IMU and video encoder architectures we utilized, which explicitly model the temporal properties of timeseries modalities. Third, we found that OOD performance of the proposed cross-modal pretraining approach is at least as good as in-distribution performance when pretraining is performed on a large IMU dataset, such as Ego4D ($>$800 hours of data). This result indicates that pretraining on large-scale datasets enables learning of generalizable features that transfer better to downstream tasks, even on different datasets, compared to when pretraining and evaluation are performed on the same smaller dataset. 

\noindent\textbf{Limitations \& future directions}: There are some limitations in our current study we hope to address in future work. First, although we utilized multiple datasets spanning a range of distribution shifts to pretrain and evaluate the proposed approach, our OOD evaluation is still fairly limited. The real-world distribution shifts may arise from a variety of sources, including hardware, subjects, diseases, sensor placement, and activities. We hope to evaluate our approach under these shifts in future work, potentially using larger study cohorts. 
Second, we treated chest-worn and head-worn IMU sensor data as similar domains. However, those minor differences could explain the reduced downstream performance in the PD dataset compared to the MMEA dataset. We believe that the pretrained IMU encoder could benefit from domain adaptation or generalization methods to improve its downstream performance on patient populations. Third, although we demonstrate improvements in zero-shot evaluations for the proposed approach over alternatives, there is still significant room for improving zero-shot performance on real-world data, especially in populations with movement disorders. Our future efforts will focus on addressing these limitations by including additional modalities or by explicitly modeling distribution shifts. 

%% file: acknowledgment.tex
\section{Acknowledgments}
This study was supported in part by the Minnesota Robotics Institute (MnRI) and the National Science Foundation grant IIS-2337909.